\documentclass{svproc}
\usepackage{url}

\usepackage[backend=biber,
            maxbibnames=99,
            natbib=true,
            sorting=anyt,%anyt
            style=numeric,%authoryear,
            citestyle=numeric-comp,%authoryear,
            url=false,
            eprint=false,
            isbn=false,
            doi=false,
            autocite=inline]{biblatex}
\usepackage{amsmath}
\newcommand{\R}{\mathcal{R}}
\newcommand{\N}{\mathcal{N}}
\newcommand{\id}{\,\mathrm{d}}

\newcommand{\der}[0]{\mathrm{d}}
\newcommand{\E}[1]{\mathbf{E}\left[ #1  \right]}
\newcommand{\Pas}{$\mathrm{P}$-a.s.}
\newcommand{\vecdiv}[0]{\operatorname{\overrightarrow{\operatorname{div}}}}
\newcommand{\grad}[0]{\operatorname{grad}}
\DeclareMathOperator*{\Tr}{Tr}
\DeclareMathOperator{\Hess}{Hess}
\DeclareMathOperator{\diver}{div}

\usepackage{array}
\usepackage{textgreek}
\usepackage{subcaption}
\usepackage{graphicx}
\usepackage{float}
\usepackage{tikz}
    \usetikzlibrary{cd, positioning, calc, fit}
    \usetikzlibrary{decorations.pathreplacing,calligraphy}
    \tikzstyle{every picture}+=[remember picture]
\usepackage{tabularx}
\usepackage{multirow}
\usepackage{tablefootnote}
\usepackage{algpseudocode, algorithm, algorithmicx}

\newdimen\figrasterwd
\figrasterwd\textwidth

\begin{document}
\mainmatter              % start of a contribution
\title{Deep Learning for the Benes Filter}
\titlerunning{Deep Learning for the Benes Filter}  % abbreviated title (for running head)
\author{Alexander Lobbe}
\authorrunning{Alexander Lobbe} % abbreviated author list (for running head)
%Craig Chambers, Kim B. Bruce, and Elisa Bertino}
%
\institute{Imperial College London, Department of Mathematics, London SW7 2AZ, United~Kingdom\\
\email{alex.lobbe@imperial.ac.uk}}
\maketitle              % typeset the title of the contribution

\begin{abstract}
The Benes filter is a well-known continuous-time stochastic filtering model in one dimension that has the advantage of being explicitly solvable. 
From an evolution equation point of view, the Benes filter is also the solution of the filtering equations given a particular set of coefficient functions.
In general, the filtering stochastic partial differential equations (SPDE) arise as the evolution equations for the conditional distribution of an underlying signal given partial, and possibly noisy, observations.
Their numerical approximation presents a central issue for theoreticians and practitioners alike, who are actively seeking accurate and fast methods, especially for such high-dimensional settings as numerical weather prediction, for example.
In this paper we present a brief study of a new numerical method based on the mesh-free neural network representation of the density of the solution of the Benes model achieved by deep learning. Based on the classical SPDE splitting method, our algorithm includes a recursive normalisation procedure to recover the normalised conditional distribution of the signal process. Within the analytically tractable setting of the Benes filter, we discuss the role of nonlinearity in the filtering model equations for the choice of the domain of the neural network. Further we present the first study of the neural network method with an adaptive domain for the Benes model.
% We would like to encourage you to list your keywords within
% the abstract section using the \keywords{...} command.
\end{abstract}
\section{Introduction}
In this paper we present a further study of the deep learning method developed in~\cite{crisan2022application} on the example of the Benes filter.
The algorithm is derived from the splitting method for SPDEs and replaces the PDE approximation step by a neural network representation and learning algorithm. Combined with the Monte-Carlo method for the approximation of the required normalisation constant, this method becomes completely mesh free.
Furthermore, an important property of the methodology in the filtering context is the ability to iterate it over several time steps. This allows the algorithm to be run \emph{online} and to successively process observations arriving sequentially.
In~\cite{crisan2022application} it was noted that a possible extension of the approximation method would be given by an adaptive domain as the support of the neural network. We present in this work the first results obtained using an adaptive domain in the nonlinear and analytically tractable case of the Benes filter.

The paper is structured as follows.
In subsection~\ref{sec:filtering} we briefly introduce the nonlinear, continuous-time stochastic filtering framework. The setting is identical to the one assumed in~\cite{crisan2022application} and the reader may consult~\cite{bain_crisan_2008} for an in-depth treatment of stochastic filtering.
Thereafter, in subsection~\ref{sec:benes_filter_model}, we formulate the Benes filtering model we will be using in the numerical studies and state the explicit solution for the filter in this case.
Then, in subsection~\ref{sec:F_eq} we introduce the filtering equation and the classical SPDE splitting method upon which the new algorithm in~\cite{crisan2022application} was built.

Next, in section~\ref{sec:dl_algo_derivation} we present an outline of the derivation of the new methodology. For details, the reader is referred to the original article~\cite{crisan2022application}. The first idea of the algorithm, presented in subsection~\ref{sec:fkp} is to reformulate the solution of the PDE for the density of the unnormalised filter as an expected value by the Feynman-Kac formula, based on an auxiliary diffusion process derived from the model equations.
Moreover, in subsection~\ref{sec:NN_model} we briefly specify the neural network parameters used in the method, as well as the employed loss-function. The theoretical part of the paper is concluded with  subsection~\ref{sec:mc_correction} where we show how to normalise the obtained neural network from the prediction step using Monte-Carlo approximation for linear sensor functions.

Section~\ref{sec:num_res} contains the detailed parameter values and results of the numerical studies that we performed.
The first result in this work is presented in subsection~\ref{sec:1d-benes_res} and is a simulation of the Benes filter using the deep learning method over a larger domain, as well as longer time interval than in the paper~\cite{crisan2022application}. Here, we have not employed any domain adaptation. However, the size of the domain needed to accommodate the support of the filter over a longer time period was estimated using the solution of the Benes model. This is necessary, as the nonlinearity of the Benes model makes it difficult to know the evolution of the posterior a priori. Thus we would be requiring a much larger domain, if chosen in an ad-hoc way.
Then, in subsection~\ref{sec:benes_res_adapted} we show our results for the Benes filter using domain adaptation. The adaptation was performed using precomputed estimates of the support of the filter by again employing the solution formula for the Benes filter.

Finally, we formulate the conclusions from our experiments in section~\ref{sec:conclusion}. In short, the domain adapted method was more effective in resolving the bimodality in our study than the non-domain adapted one. However, this came at the cost of a linear trend in the error.

\subsection{Nonlinear stochastic filtering problem}\label{sec:filtering}
The stochastic filtering framework consists of a pair of stochastic processes $(X,Y)$ on a probability space $(\Omega, \mathcal{F}, \mathrm{P})$ with a normal filtration $\,(\mathcal{F}_t)_{t\geq0}$ modelled, \Pas, as
\begin{equation}\label{eq:signal}
    X_t = X_0 + \int_0^t f(X_s) \id s + \int_0^t \sigma(X_s) \id V_s \;,
\end{equation}
and
\begin{equation}\label{eq:observation}
    Y_t = \int_0^t h(X_s) \id s + W_t \;.
\end{equation}
Here, the time parameter is $t\in[0,\infty)$, $d,p\in\N$ and $f: \R^d \rightarrow \R^d$ and
$\sigma: \R^d \rightarrow \R^{d \times p}$ are the drift and diffusion coefficient functions of the signal. The processes
$V$ and $W$ are $p$-- and $m$-dimensional independent, $(\mathcal{F}_t)_{t\geq0}$-adapted Brownian motions.
We call $X$ the \emph{signal process} and $Y$ the \emph{observation process}. The function $h: \R^{d} \rightarrow \R^{m}$ is often called the \emph{sensor function}, or \emph{link function}, because it models the possibly nonlinear connection of the signal and observation processes.

Further, consider the \emph{observation filtration} $(\mathcal{Y}_t)_{t\geq0}$ given as
\begin{equation*}
    \mathcal{Y}_t = {\sigma}(Y_s, s\in[0,t]) \vee \mathcal{N}
    \hspace{10pt}\text{ and }\hspace{10pt} \mathcal{Y} =
    \sigma\left(\bigcup_{t \in [0,\infty)} \mathcal{Y}_t\right),
\end{equation*}
where $\mathcal{N}$ are the $ \mathrm{P}$-nullsets of
$\mathcal{F}$. The aim of nonlinear filtering is to compute the probability measure valued $(\mathcal{Y}_t)_{t\geq 0}$-adapted stochastic 
process $\pi$ that is defined by the requirement that for all bounded measurable test functions
$\varphi: \R^d \to \R$ and $t\in [0,\infty)$ we have \Pas{} that
\begin{equation*}
    \pi_t\varphi = \E{ \varphi(X_t) \left| \mathcal{Y}_t \right. }.
\end{equation*}
We call $\pi$ the \emph{filter}.

Furthermore, let the  process $Z$ be defined such that for all $t\in[0,\infty)$,
\begin{equation*}
Z_t = \exp\{-\int_0^t h(X_s)\id W_s - \frac{1}{2} \int_0^t h(X_s)^2 \id s\}.
\end{equation*}
Then, assumimg that
\begin{equation*}
    \E{\int_0^t h(X_s)^2 \id s} < \infty \;
    \text{ and } \; \E{\int_0^t Z_s h(X_s)^2 \id s} < \infty,
\end{equation*}
we have that $Z$ is an 
$(\mathcal{F}_t)_{t\geq0}$-martingale and by the change of measure (for details, see~\cite{bain_crisan_2008}) given by 
$\left. \frac{\id \tilde{\mathrm{P}}^t}
{\id \mathrm{P}}\right|_{\mathcal{F}_t} = Z_t$, $t\geq 0$,
the processes $X$ and $Y$ are independent under $\tilde{\mathrm{P}}$ and $Y$ is a $\tilde{\mathrm{P}}$-Brownian motion. Here, $\tilde{\mathrm{P}}$
is the consistent measure defined on $\bigcup_{t\in [0,\infty)}\mathcal{F}_t$.
Finally, under $\tilde{\mathrm{P}}$, we can define the measure valued stochastic process $\rho$
by the requirement that for all bounded measurable functions
$\varphi: \R^d\to\R$ and $t\in [0,\infty)$ we have \Pas{} that
\begin{equation}\label{eq:rho}
    \rho_t\varphi = \E{ \left.\varphi(X_t) 
    \exp\{\int_0^t h(X_s)\id Y_s - \frac{1}{2} \int_0^t h(X_s)^2 \id s\}
    \right| \mathcal{Y}_t }.
\end{equation}
The Kallianpur-Striebel formula (see~\cite{bain_crisan_2008}) justifies the terminology to call $\rho$ the \emph{unnormalised filter}.

\subsection{The Benes filtering model}\label{sec:benes_filter_model}

The Benes filter is is a one-dimensional nonlinear model and is used as a benchmark in the numerical studies below. As we show below, it is one of the rare cases of explicitly solvable continuous-time stochastic filtering models.
Here, we are considering a special case of the more general class of Benes filters, presented, for example, in~\cite[Chapter 6.1]{bain_crisan_2008}.

The signal is given by the coefficient functions
\begin{equation*}
    f(x) = \alpha\sigma \tanh (\beta+\alpha x/ \sigma) \;\text{ and } \;
    \sigma(x) \equiv \sigma \in \R,
\end{equation*}
where $\alpha, \beta \in \R$
and the observation is given by the affine-linear sensor function
\begin{equation*}
    h(x) = h_1 x + h_2,
\end{equation*}
with $h_1, h_2 \in \R$. 
The density $p_B$ of the filter solving the Benes model is then given by two weighted Gaussians as
\begin{equation}
    p_B(z) = w^{+}\Phi(\mu_t^{+}, \nu_t)(z) + w^{-}\Phi(\mu_t^{-}, \nu_t)(z),
\end{equation}
where $\mu_t^{\pm} = M_t^{\pm}/(2v_t)$, $\nu_t= 1/(2v_t)$, and 
\begin{equation*}
    w^{\pm} = \frac{\exp ( (M_t^{\pm})^2/(4v_t))}{\exp ( (M_t^{+})^2/(4v_t))\exp ( (M_t^{-})^2/(4v_t))}
\end{equation*}
with
\begin{equation*}
    M_t^{\pm} = \pm\frac{\alpha}{\sigma} + h_1\int_0^t \frac{\sinh(s\zeta\sigma )}{\sinh(t\zeta\sigma)}\id Y_s +\frac{h_2+h_1x_0}{\sigma\sinh(t\zeta\sigma)} - \frac{h_2}{\sigma}\coth(t\zeta\sigma),
\end{equation*}
$v_t = {h_1}\coth(t\zeta\sigma)/{2\sigma}$,
and $\zeta = \sqrt{\alpha^2/\sigma^2 + h_1^2}$.

\subsection{Filtering equation and general splitting method}\label{sec:F_eq}
Note that under the conditions given in~\cite{crisan2022application}, $X$ admits the infinitesimal generator
$A: \mathcal{D}(A) \rightarrow B(\R^{d})$ given,
for all $\varphi \in \mathcal{D}(A)$, by
\begin{equation}\label{eq:A}
    A\varphi = \langle f, \nabla \varphi \rangle + \Tr(a\Hess\varphi),
\end{equation}
where $\mathcal{D}(A)$ denotes the domain of the differential operator $A$
and $a = \frac{1}{2}\sigma\sigma^\prime$.

It is well-known (see, e.g., \cite{bain_crisan_2008}), 
that the unnormalised filter $\rho$ satisfies the \emph{filtering equation}, i.e. for all $t\geq 0$, we have $\tilde{\mathrm{P}}$-a.s. that
\begin{equation}
    \label{eq:Zakai_rho}
    \rho_t(\varphi) = \pi_0(\varphi) +
        \int_0^t \rho_s(A\varphi)\id s +
        \int_0^t \rho_s(\varphi h^\prime) \id Y_s.
\end{equation}

The classical splitting method for the filtering equation is given in~\cite{cai1995adaptive} and
seeks to approximate
the following SPDE for the density $p_t$ of the unnormalised filter given, for all $t\geq0$, $x\in\R^d$, 
and \Pas{} as
\begin{equation*}
    p_t(x) = p_0 (x) + \int_0^t A^* p_s(x)\id s + 
    \int_0^t h^\prime(x) p_s(x)  \id Y_s
\end{equation*}
and relies on the splitting-up algorithm described in \cite{le1989time}
and \cite{legland1992splitting}. Here $A^*$ is the formal adjoint of the infinitesimal generator $A$ of the signal process $X$.

We summarise the splitting-up method below in Note~\ref{meth:PDE}.

\begin{note}\label{meth:PDE}
The splitting method for the filtering problem is defined by iterating the steps below with initial density
${p}^{0}(\cdot)= p_0(\cdot)$:
\begin{enumerate}
    \item\label{predictor}\emph{(Prediction)}
    Compute an approximation $\tilde{p}^n$ of the solution to
    \begin{equation}\label{eq:num_IVP}
        \begin{aligned}
        \frac{\partial q^n}{\partial t} (t,z) &= A^* q^n (t,z), 
        &\; & (t,z)\in (t_{n-1},t_n]\times\R^d,\\
            q^n(0,z) &= {p}^{n-1}(z), &\;    & z\in \R^d,
        \end{aligned}
    \end{equation}
    at time $t_n$ and
    \item\emph{(Normalisation)}
    Compute the normalisation constant with
    $z_n = (Y_{t_n}-Y_{t_{n-1}}) / (t_n-t_{n-1})$
    and the function
    \begin{equation*}
        \R^d\ni z\mapsto\xi_n(z) = \exp \left( -\frac{t_n-t_{n-1}}{2} 
        ||z_n - h(z)||^2 \right),
    \end{equation*}
    so that we can set,
    \begin{equation*}
        p^n(z) = \frac{1}{C_n} \xi_n(z) \tilde{p}^n(z); \; z\in\R^d,
    \end{equation*}
    where $C_n =
    \int_{\R^d} \xi_n(z)\tilde{p}^n(z) \id z $.
\end{enumerate}
\end{note}
The deep learning method studied below replaces the predictor step of the splitting method above by a deep neural network approximation algorithm to avoid an explicit space discretisation.
This is achieved by representing each $\tilde{p}^n(z)$ by a feed-forward
neural network and approximating the initial value problem \eqref{eq:num_IVP}
based on its stochastic representation using a sampling procedure.
The normalisation step may then be computed either using quadrature, or, to preserve the mesh-free characteristic, by Monte-Carlo approximation.

\section{Derivation and outline of the deep learning algorithm}\label{sec:dl_algo_derivation}

Here, we present a concise version of the derivation laid out in detail in~\cite{crisan2022application}.

\subsection{Feynman-Kac representation}\label{sec:fkp}
Assuming sufficient differentiability of the coefficient functions, the operator $A^*$ may be expanded such that for all $\varphi \in C_c^\infty(\R^d,\R)$ we have
    \begin{equation}
    \label{eq:op_expansion}
        A^* \varphi = \Tr(a\Hess \varphi) +
            \langle 2\vecdiv(a)-f, \grad \varphi \rangle +
            \diver(\vecdiv(a) - f)\varphi.
    \end{equation}

Subtracting the zero-order term from \eqref{eq:op_expansion}, we obtain an operator that generates the auxiliary diffussion process, denoted $\hat{X}$, which is instrumental in the deep learning method.
\begin{definition}\label{def:A_hat}
    Define the partial differential operator 
    $\hat{A}:C_c^\infty(\R^d,\R)\rightarrow C_b(\R^d,\R)$ such that
    for all $\varphi \in C_c^\infty(\R^d,\R)$,
    \begin{equation*}
        \hat{A} \varphi = \Tr(a\Hess \varphi) +
            \langle 2\vecdiv(a)-f, \grad \varphi \rangle
    \end{equation*}
    and the function $r:\R^d\rightarrow \R$ such that for all
    $x\in \R^d$,
    \begin{equation*}
        r(x) = \diver(\vecdiv(a)-f)(x).
    \end{equation*}
\end{definition}

\begin{lemma}\label{prop:diffusion}
    For all $x\in \R^d$ the operator $\hat{A}$ defined in
    Definition~\ref{def:A_hat} is the infinitesimal generator
    of the Itô diffusion $\hat{X}:[0,\infty)\times \Omega \rightarrow \R^d$
    given, for all $t\geq0$ and \Pas{} by
    \begin{equation*}
        \hat{X}_t =x + \int_0^t b(\hat{X}_s) \der s +
            \int_0^t \sigma(\hat{X}_s) \der \hat{W}_s,
    \end{equation*}
    where $\hat{W}:[0,\infty)\times \Omega \rightarrow \R^d$
    is a $d$-dimensional Brownian motion and
    $b:\R^d \rightarrow \R^d$ is the function
    \begin{equation*}
        b = 2\vecdiv(a) - f.
    \end{equation*}
\end{lemma}

From the well-known Feynman-Kac formula (see Karatzas and Shreve~\cite[Chapter 5, Theorem 7.6]{KaratzasShreve1998})  we can deduce the
Corollary~\ref{thm:initialValuePDE} below for the
initial value problem.

\begin{corollary}\label{thm:initialValuePDE}
Let $d\in\N$, $T>0$, $k\in C(\R^d, [0,\infty))$,
    let $\hat{A}$ be the operator defined in
    Definition~\ref{def:A_hat},
    and let $\psi:\R^d \rightarrow \R$ be an at most polynomially growing
    function.
    Suppose that $u \in C_b^{1,2}((0,T]\times \R^d,\R)$ satisfies the Cauchy 
    problem
    \begin{equation}\label{eqn:initialValuePDE}
    \begin{aligned}
         \frac{\partial u}{\partial t}(t,x) + k(x)u(t,x) &= \hat{A}u(t,x),  &\; 
         &(t,x)\in (0,T] \times \R^d,\\
        u(0,x) &= \psi(x), &\; &x\in \R^d.
    \end{aligned}
    \end{equation}
    Then, for all $(t,x) \in (0,T]\times \R^d$, we have that
    \begin{equation*}
        u(t,x) = \E{\left. \psi(\hat{X}_t)\exp\left(-\int_0^t k(\hat{X}_\tau)
        \id\tau\right) \right| \hat{X}_0 = x},
    \end{equation*}
    where $\hat{X}$ is the diffusion generated by $\hat{A}$.
\end{corollary}

Recall that our aim is to approximate the Fokker-Planck equation \eqref{eq:num_IVP}. Written in the form as in Corollary~\ref{thm:initialValuePDE}, \eqref{eq:num_IVP} reads as
\begin{equation*}
    \begin{aligned}
    \frac{\partial q^n}{\partial t} (t,z) &= \hat{A}q^n(t,z) + r(z)q^n(t,z), 
    &\; & (t,z)\in (t_{n-1},t_n]\times\R^d,\\
        q^n(0,z) &= {p}^{n-1}(z), &\;    & z\in \R^d.
    \end{aligned}
\end{equation*}
Thus, with $k = -r$, and assuming that $-r$ is non-negative
in \eqref{eqn:initialValuePDE}, we obtain by Corollary~\ref{thm:initialValuePDE} the representation,
for all $n\in\{1,\dots,N\}$, $t\in (t_{n-1},t_n]$, $z\in \R^d$, 
\begin{equation}\label{eq:fk_exp}
    q^n(t,z) = \E{\left. p^{n-1}(\hat{X}_t)
        \exp\left(\int_{t_{n-1}}^t r(\hat{X}_\tau) \id\tau\right) \right|
        \hat{X}_{t_{n-1}} = z}.
\end{equation}

Note that \cite[Proposition 2.4]{crisan2022application} shows that we have a feasible minimisation problem to approximate by the learning algorithm (see also~\cite[Proposition 2.7]{beck2018solving}).

\begin{remark}
For the Benes model, note that the auxiliary diffusion is given as
\begin{equation*}
    \hat{X}_t = \hat{X}_0 - \int_0^t \alpha\sigma\tanh(\beta + \alpha x / \sigma) \, \der s +
        \int_0^t \sigma \,\der \hat{W}_s,
\end{equation*}
and the coefficient
\begin{equation*}
    r(x) = - \diver f(x) = 
        -\alpha^2 \operatorname{sech}^2(\beta + \alpha x / \sigma).
\end{equation*}
Therefore the representation of the solution to the Fokker-Planck equation~\eqref{eq:num_IVP} in the Benes case reads
\begin{equation*}
    q^n(t,z) = \E{\left. p^{n-1}(\hat{X}_t)
        \exp\left(- \int_{t_{n-1}}^t \alpha^2\operatorname{sech}^2(\beta + \alpha \hat{X}_\tau /\sigma) \,\der \tau\right) \right|
        \hat{X}_{t_{n-1}} = z}.
\end{equation*}
\end{remark}

\subsection{Neural network model for the prediction step}\label{sec:NN_model}

To solve the Fokker-Planck equation over a rectangular domain $\Omega_d = [\alpha_1, \beta_1]\times\dots\times[\alpha_d,\beta_d]$, we employ the sampling based deep learning method from~\cite{beck2018solving}.
Using the representation~\eqref{eq:fk_exp}, the solution of the Fokker-Planck equation is reformulated into an optimisation problem over function space given in \cite[Proposition 2.4]{crisan2022application}. This in turn yields the loss functions for the learning algorithm. Writing $\hat{\mathrm{X}}^\xi$ for the auxiliary diffusion with $\text{Unif}(\Omega_d)$-random initial value $\xi$, the optimisation problem is approximated by the optimisation
\begin{equation*}
    \inf_{\theta \in \R^{\sum_{i=2}^L l_{i-1}l_i+l_i}} \E{\left| \psi(\hat{\mathrm{X}}_T^\xi)
        \exp\left(-\int_0^T k(\hat{\mathrm{X}}_\tau^\xi) \id\tau\right) -
        \mathcal{NN}_\theta(\xi)\right|^2}
\end{equation*}
where the solution of the PDE is represented by a neural network $\mathcal{NN}_\theta$ and the infinite-dimensional function space has been parametrised by $\theta$. Here, $L$ denotes the depth of the neural net, and the parameters $l_i$ are the respective layer widths. Further details can be found in~\cite{crisan2022application}. A comprehensive textbook on deep learning is~\cite{goodfellow2016deep}.
We apply a modified gradient descent method, called ADAM~\cite{kingma2014adam}, to determine the parameters in the model by minimising the \emph{loss function}
\begin{multline*}
\mathcal{L}(\theta ; \{\xi^i, \{\hat{\mathrm{X}}_{\tau_j}^{\xi,i}\}_{j=0}^J \}_{i=1}^{N_b}) =\\
    \frac{1}{N_b} \sum_{i=1}^{N_b} \left| 
        \psi(\hat{\mathrm{X}}_T^{\xi,i})
        \exp(- \sum_{j=0}^{J-1} k(\hat{\mathrm{X}}_{\tau_j}^{\xi,i}) (\tau_{j+1}-\tau_j)) -
        \mathcal{NN}_\theta(\xi^i)\right|^2,
\end{multline*}
where $N_b$ is the batch size and $ \{\xi^i, \{\hat{\mathrm{X}}_{\tau_j}^{\xi,i}\}_{j=0}^J \}_{i=1}^{N_b}$ is a training batch of independent identically distributed realisations $\xi^i$ of $\xi \sim \mathcal{U}(\Omega_d)$ and $\{\hat{\mathrm{X}}_{\tau_j}^{\xi,i}\}_{j=0}^J$ the approximate i.i.d.{} realisations of sample paths of the auxiliary diffusion started at $\xi^i$ over the time-grid $\tau_0=0<\tau_1<\cdots<\tau_{J-1}<\tau_J=T$. For the approximation of the sample paths of the diffusion we use the Euler-Maruyama method~\cite{KloedenPlaten1992}.
Additionally, we augment the loss $\mathcal{L}$ by an additional term to encourage the positivity of the neural network. Thus, in practice, we use the loss
\begin{equation*}
    \tilde{\mathcal{L}}(\theta ;  \{\xi^i, \{\hat{\mathrm{X}}_{\tau_j}^i\}_{j=0}^J \}_{i=1}^{N_b}) =
    {\mathcal{L}}(\theta ;  \{\xi^i, \{\hat{\mathrm{X}}_{\tau_j}^i\}_{j=0}^J \}_{i=1}^{N_b}) 
    + \lambda \sum_{i=1}^{N_b} \max\{0,\mathcal{NN}_\theta(\xi^i)\}
\end{equation*}
with the hyperparameter $\lambda$ to be chosen.

Thus, in the notation of subsection~\ref{sec:F_eq}
we replace the Fokker-Planck solution by a neural network model,
i.e. we \emph{postulate} a neural network model
\begin{equation*}
    \tilde{p}_n(z) = \mathcal{NN}(z),
\end{equation*}
with support on $\Omega_d$. Therefore we require the a priori chosen domain to capture most of the mass of the probability distribution it is approximating.

\subsection{Monte-Carlo normalisation step}\label{sec:mc_correction}
We then realise the normalisation step via Monte-Carlo sampling over the bounded rectangular domain $\Omega_d$ to approximate the integral
\begin{equation}\label{eq:MC_int}
     \int_{\R^d} \xi_n(z)\mathcal{NN}(z) \id z = \int_{\Omega_d} \exp \left( -\frac{t_n-t_{n-1}}{2} 
        ||z_n - h(z)||^2 \right) \mathcal{NN}(z) \id z,
\end{equation}
where, as defined earlier, $z_n = \frac{1}{t_n-t_{n-1}}(Y_{t_n}-Y_{t_{n-1}})$. Note that, since $\Omega_d$ is the support of the neural network $\mathcal{NN}$, the right-hand side above is indeed identical to the integral over the whole space.

The sensor function in the Benes model is given by $h(x) = h_1 x + h_2$.
Then, the likelihood function becomes
\begin{equation*}
    \xi_n(z) = \frac{\sqrt{2\pi}}{\sqrt{(t_n-t_{n-1})h_1^2}} \mathcal{N}_{\text{pdf}}\left(\frac{z_n - h_2}{h_1}, \frac{1}{(t_n-t_{n-1})h_1^2}\right)(z),
\end{equation*}
where $\mathcal{N}_{\text{pdf}}(\mu,\sigma^2)$ denotes the probability density function of a normal distribution with mean $\mu$ and variance $\sigma^2$. Therefore, we can write the integral \eqref{eq:MC_int} as
\begin{equation*}
    \frac{\sqrt{2\pi}}{\sqrt{(t_n-t_{n-1})h_1^2}} \mathbf{E}_{Z}[\mathcal{NN}(Z)]; \qquad  \qquad Z \sim \mathcal{N}\left(\frac{z_n - h_2}{h_1}, \frac{1}{(t_n-t_{n-1})h_1^2}\right).
\end{equation*}
This is an implementable method to compute the normalisation constant $C_n$. Thus, we can express the approximate posterior density as
\begin{equation*}
        p^n(z) = \frac{1}{C_n} \xi_n(z) \tilde{p}^n(z).
\end{equation*}
Therefore, the methodology is fully recursive and can be applied sequentially.

\begin{remark}\label{rem:mc_sampling}
In low-dimensions, the usage of the Monte-Carlo method to perform the normalisation is optional, since efficient quadrature methods are an alternative.
We chose the sampling based method to preserve the grid-free nature of the algorithm.
\end{remark}

\section{Numerical results for the Benes filter}\label{sec:num_res}

The neural network architecture for all our experiments below is a feed-forward fully connected neural network with a one-dimensional input layer, two hidden layers with a layer width of $51$ neurons each and batch-normalisation, and an output layer of dimension one (a detailed illustration can be found in~\cite{crisan2022application}). For the optimisation algorithm we chose the ADAM optimiser and performed the training over $6002$ epochs with a batch size of $600$ samples. The initial signal and observation values are $x_0=y_0=0$ and the coefficients of the Benes model were chosen as $\alpha = 3$, $\beta=0$, $\sigma=0.5$, $h_1=3$, $h_2=0$, and timestep $\Delta t = 0.1$ over $N=40$ steps. The initial condition is a Gaussian density with mean $0$ and standard deviation $0.001$. The posterior was calculated over the domain $[-9,2.5]$. The domain boundaries were pre-estimated using a simulation of the exact Benes filter with fixed random seed. In the case of the domain adaptation we used the precomputed evolutions from the true solution to estimate the support of the posterior and set a fixed domain adaptation schedule. The spatial resolution is $1000$ uniformly spaced values in the domain of definition of the neural network. At each time step, the training of the network consumes $6002\cdot600=3,601,200$ Monte-Carlo samples. Additionally we employ a piecewise constant learning rate schedule $lr(epoch) = 10^{-(2+epoch\bmod{2001})}$ and the normalisation constant is computed using $10^7$ samples each timestep. The regularising parameter $\lambda = 1$.

\subsection{No domain adaptation}\label{sec:1d-benes_res}
\begin{figure}[!ht]
  \centering
  \parbox{\figrasterwd}{
    \parbox{.55\figrasterwd}{%
      \subcaptionbox{}{\includegraphics[width=1.2\hsize]{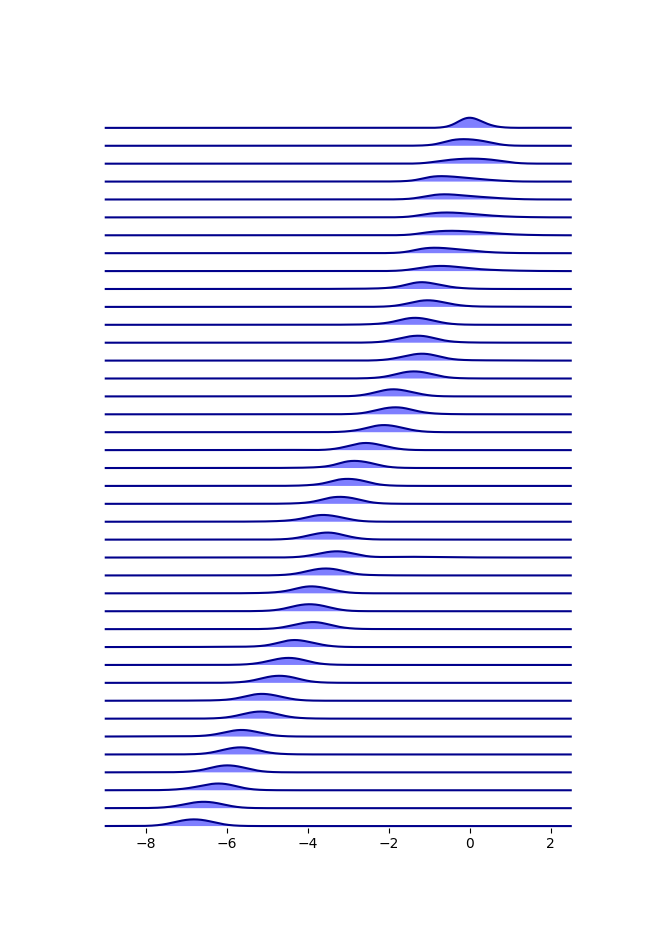}}
    }
    \hskip1em
    \parbox{.35\figrasterwd}{%
      \subcaptionbox{}{\includegraphics[width=\hsize]{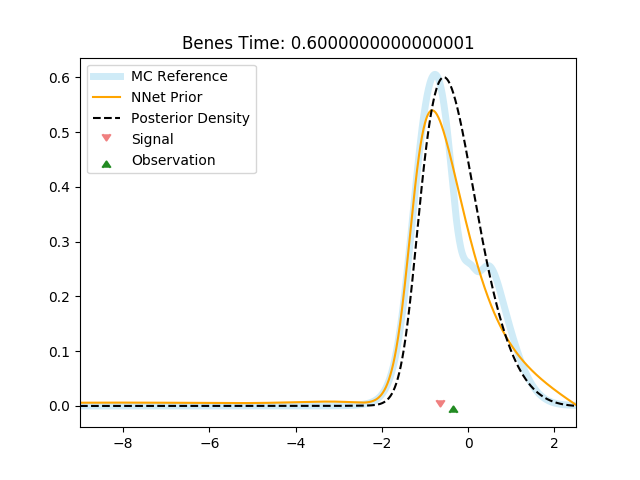}}
      \vskip1em
      \subcaptionbox{}{\includegraphics[width=\hsize]{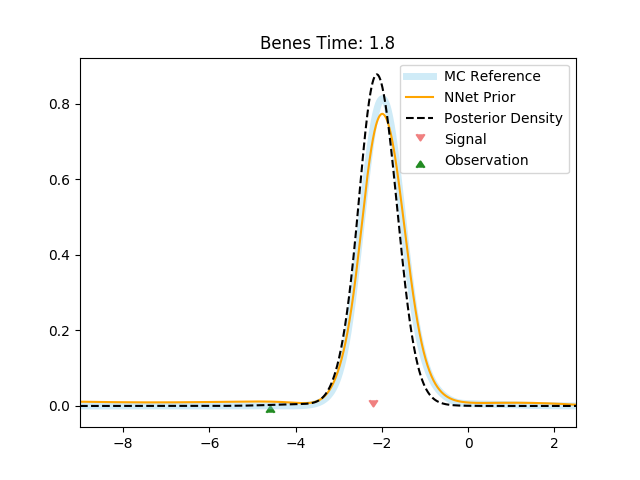}}
      \vskip1em
      \subcaptionbox{}{\includegraphics[width=\hsize]{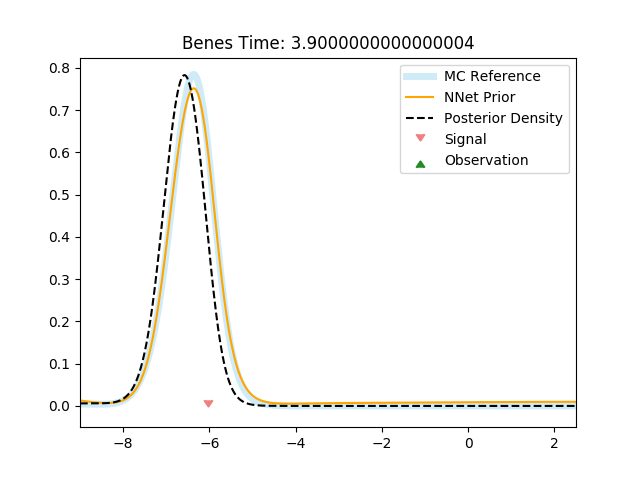}} 
    }
  }
  \caption{Results of the combined splitting-up/machine-learning approximation applied iteratively to the Benes filtering problem (no domain adaptation). (a) The full evolution of the estimated posterior distribution produced by our method, plotted at all intermediate timesteps. (b)-(d) Snapshots of the approximation at times, $t=0.6$, $t=1.8$, and $t=3.9$. The black dotted line in each graph shows the estimated posterior, the yellow line the prior estimate represented by the neural network, and the light-blue shaded line shows the Monte-Carlo reference solution for the prior.}
\label{fig:1d_ben_graphs}
\end{figure}

\begin{figure}
     \centering
     \begin{subfigure}[b]{0.48\textwidth}
         \centering
         \includegraphics[width=\textwidth]{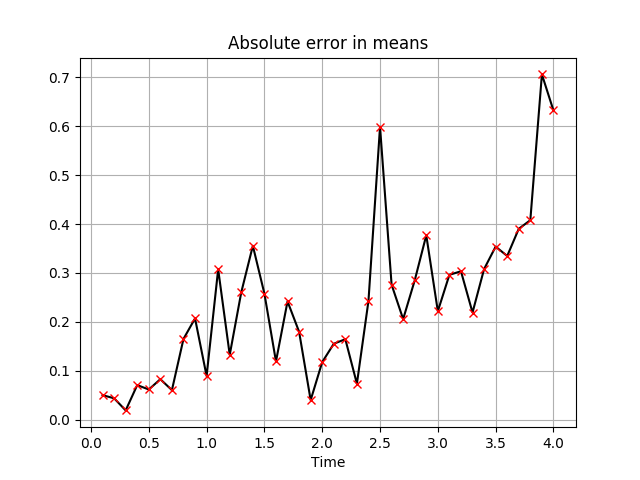}
         \caption{}
         \label{fig:y56367547srewgdsrt x}
     \end{subfigure}
     \hfill
     \begin{subfigure}[b]{0.48\textwidth}
         \centering
         \includegraphics[width=\textwidth]{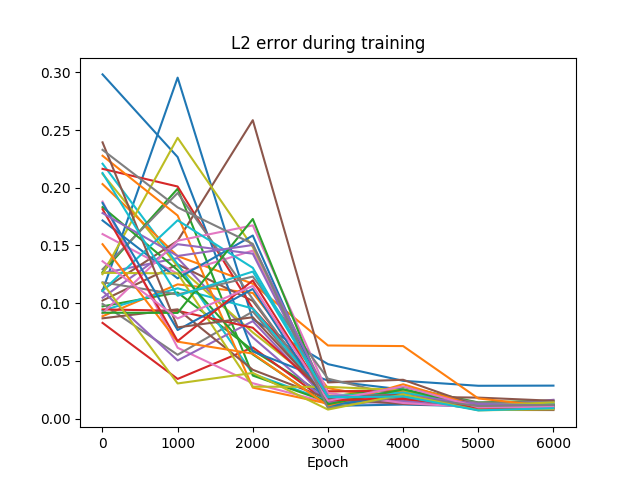}
         \caption{}
         \label{fig:5345633gdftwerg}
     \end{subfigure}
     \newline
     \begin{subfigure}[b]{0.48\textwidth}
         \centering
         \includegraphics[width=\textwidth]{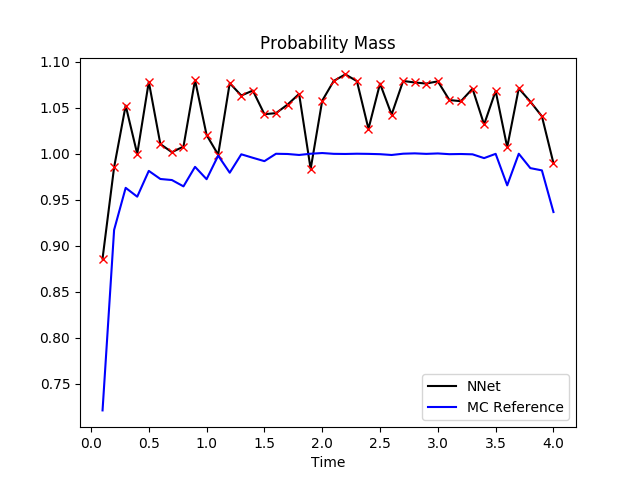}
         \caption{}
         \label{fig:5435346xhtyrtewfrews}
     \end{subfigure}
     \hfill
     \begin{subfigure}[b]{0.48\textwidth}
         \centering
         \includegraphics[width=\textwidth]{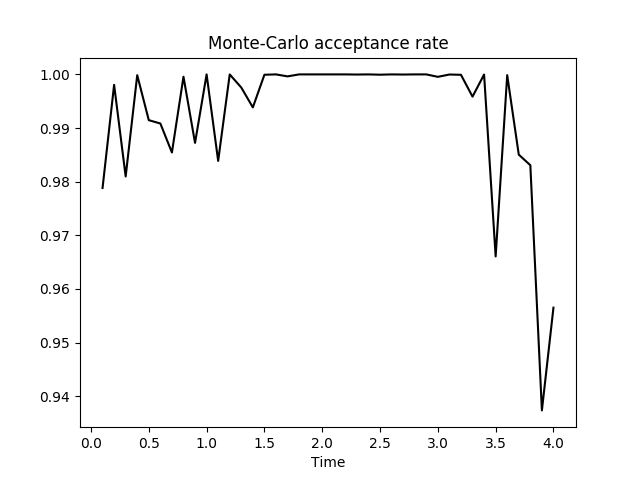}
         \caption{}
         \label{fig:t54643trhytrh42f}
     \end{subfigure}
        \caption{Error and diagnostics for the Benes filter (no domain adaptation). (a) Absolute error in means between the approximated distribution and the exact solution. (b) $L_2$ error of the neural network during training with respect to the Monte-Carlo reference solution. (c) Probability mass of the neural network prior. (d) Monte-Carlo acceptance rate.}
        \label{fig:1d_ben_graphserr}
\end{figure}

Figure~\ref{fig:1d_ben_graphs} shows the plots for the Benes filter without domain adaptation. In Figure~\ref{fig:1d_ben_graphs}(a) we observe the drift of the posterior toward the left edge of the domain. The initial bimodality, reflecting the uncertainty due to few observed values, quickly resolves and the approximate posterior tracks the signal within the domain. In~ Figure~\ref{fig:1d_ben_graphs}(b) the bimodality is mostly visible in the Monte-Carlo prior and smoothed out by the neural network. Figure~\ref{fig:1d_ben_graphs}(c)+(d) show snapshots of the progression of the filter.
The absolute error in means with respect to the Benes reference solution is plotted in Figure~\ref{fig:1d_ben_graphserr}(a) and shows that as the posterior reaches the left domain boundary, the error increases. This is reflected as well in the drop of probability mass, Figure~\ref{fig:1d_ben_graphserr}(c), and Monte-Carlo acceptance rate, Figure~\ref{fig:1d_ben_graphserr}(d) at later times. It is not clear from Figure~\ref{fig:1d_ben_graphserr}(a) if there is a trend in the error. Further experiments need to be performed to check this hypothesis.
Figure~\ref{fig:1d_ben_graphserr}(b) shows that the neural net training consistently succeeds as measured by the $L_2$ distance between the Monte-Carlo reference prior and the neural net prior.

\subsection{With domain adaptation}\label{sec:benes_res_adapted}
\begin{figure}[!ht]
  \centering
  \parbox{\figrasterwd}{
    \parbox{.55\figrasterwd}{%
      \subcaptionbox{}{\includegraphics[width=1.2\hsize]{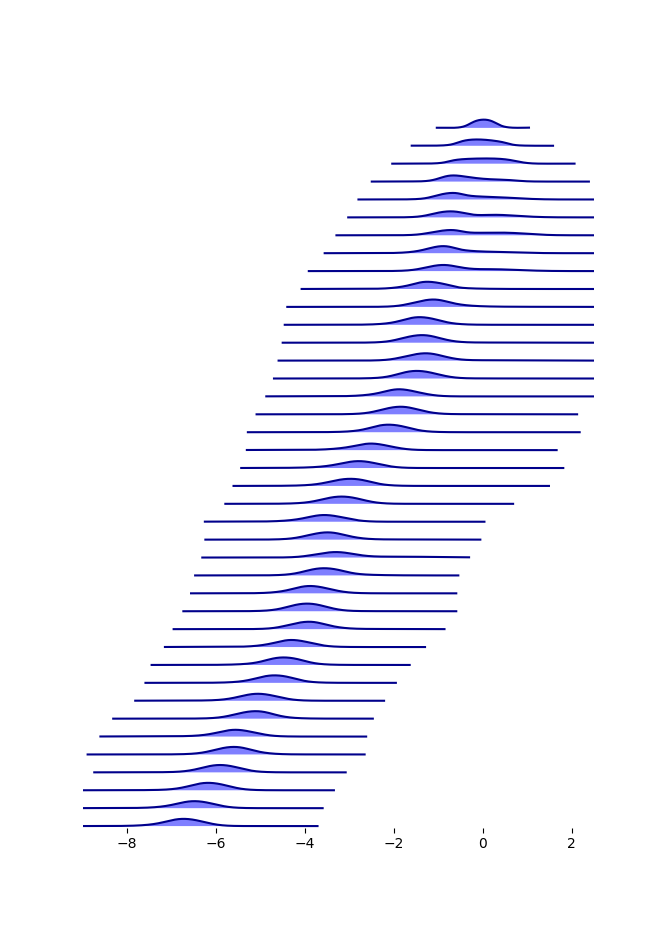}}
    }
    \hskip1em
    \parbox{.35\figrasterwd}{%
      \subcaptionbox{}{\includegraphics[width=\hsize]{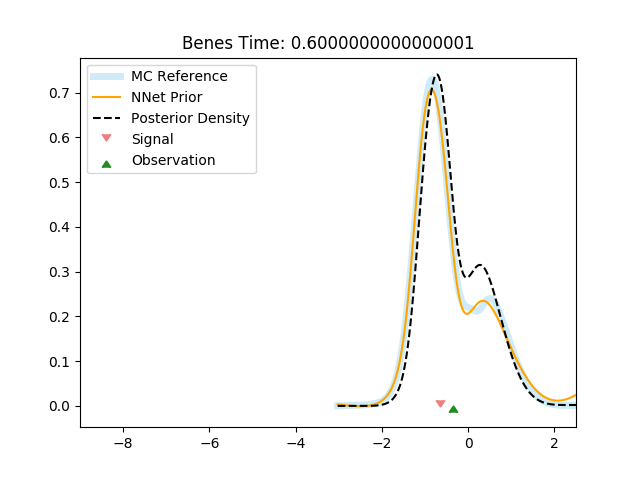}}
      \vskip1em
      \subcaptionbox{}{\includegraphics[width=\hsize]{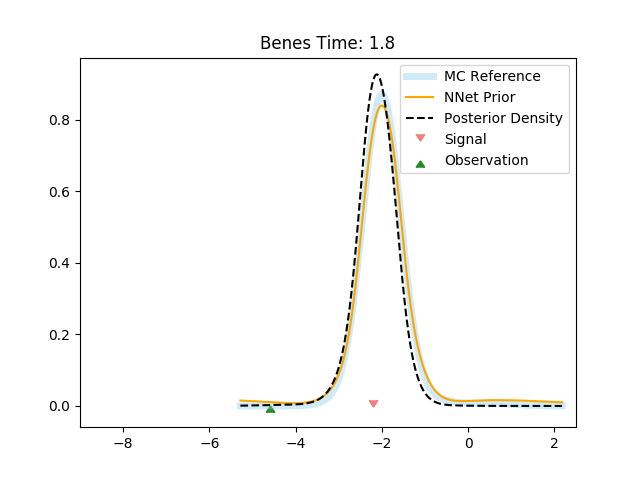}}
      \vskip1em
      \subcaptionbox{}{\includegraphics[width=\hsize]{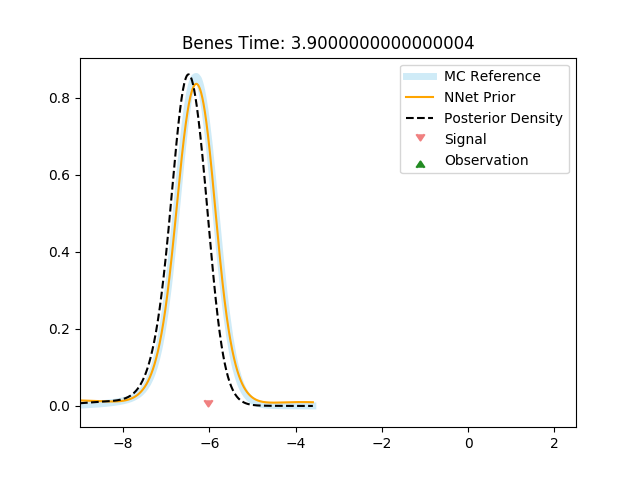}} 
    }
  }
  \caption{Results of the combined splitting-up/machine-learning approximation applied iteratively to the Benes filtering problem (with domain adaptation). (a) The full evolution of the estimated posterior distribution produced by our method, plotted at all intermediate timesteps. (b)-(d) Snapshots of the approximation at times, $t=0.6$, $t=1.8$, and $t=3.9$. The black dotted line in each graph shows the estimated posterior, the yellow line the prior estimate represented by the neural network, and the light-blue shaded line shows the Monte-Carlo reference solution for the prior.}
\label{fig:1d_ben_graphs_adapted}
\end{figure}

\begin{figure}
     \centering
     \begin{subfigure}[b]{0.48\textwidth}
         \centering
         \includegraphics[width=\textwidth]{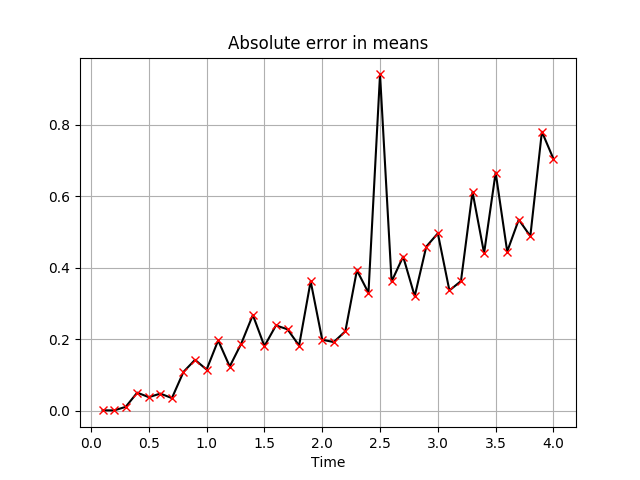}
         \caption{}
         \label{fig:y56366547srewgdsrt x}
     \end{subfigure}
     \hfill
     \begin{subfigure}[b]{0.48\textwidth}
         \centering
         \includegraphics[width=\textwidth]{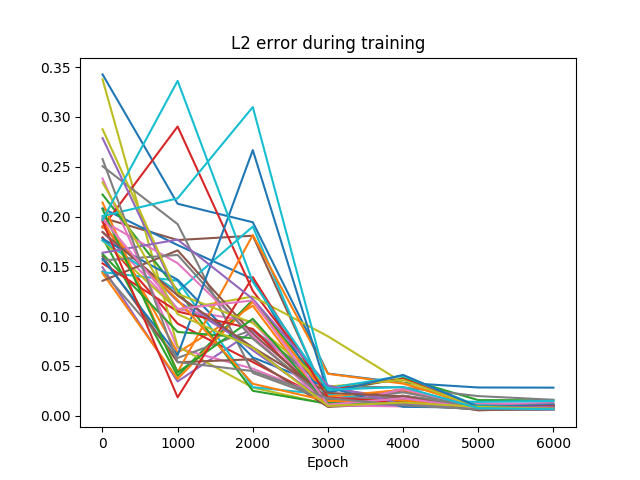}
         \caption{}
         \label{fig:5345563gdftwerg}
     \end{subfigure}
     \newline
     \begin{subfigure}[b]{0.48\textwidth}
         \centering
         \includegraphics[width=\textwidth]{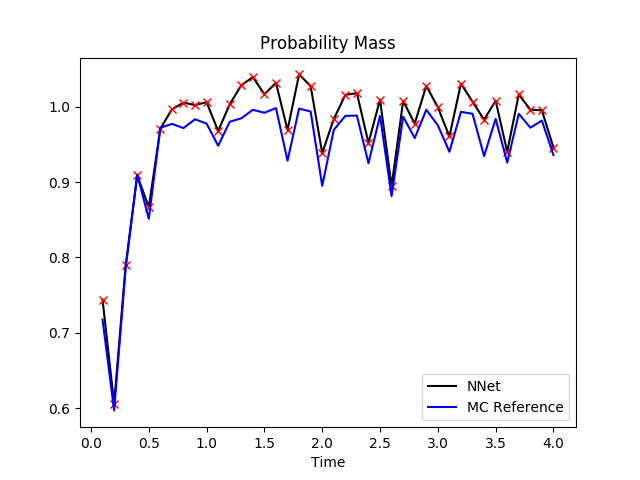}
         \caption{}
         \label{fig:543534656tewfrews}
     \end{subfigure}
     \hfill
     \begin{subfigure}[b]{0.48\textwidth}
         \centering
         \includegraphics[width=\textwidth]{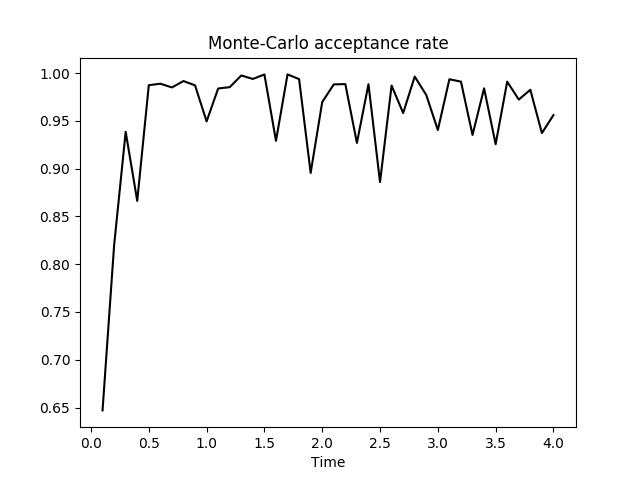}
         \caption{}
         \label{fig:t546423542f}
     \end{subfigure}
        \caption{Error and diagnostics for the Benes filter (with domain adaptation). (a) Absolute error in means between the approximated distribution and the exact solution. (b) $L_2$ error of the neural network during training with respect to the Monte-Carlo reference solution. (c) Probability mass of the neural network prior. (d) Monte-Carlo acceptance rate.}
        \label{fig:1d_ben_graphserr_adapt}
\end{figure}

Figure~\ref{fig:1d_ben_graphs_adapted} shows the plots for the Benes filter with domain adaptation. In Figure~\ref{fig:1d_ben_graphs_adapted}(a) we observe again the drift of the posterior toward the left edge of the domain. and the initial bimodality resolves. The approximate posterior tracks the signal within the domain. In~ Figure~\ref{fig:1d_ben_graphs_adapted}(b) the bimodality is visible both in the prior an the posterior network. This shows that the domain adaptation helps resolve the bimodality in the nonlinear case by increasing the spatial resolution while keeping the computational cost equal. Figure~\ref{fig:1d_ben_graphs_adapted}(c)+(d) again show snapshots of the progression of the filter.
The absolute error in means with respect to the Benes reference solution is plotted in Figure~\ref{fig:1d_ben_graphserr_adapt}(a) and shows a clear linear trend. This is an interesting phenomenon, likely due to the reduced domain size and subsequent error accumulation. The probability mass, Figure~\ref{fig:1d_ben_graphserr_adapt}(c), and Monte-Carlo acceptance rate, Figure~\ref{fig:1d_ben_graphserr_adapt}(d) are stably fluctuating.
Figure~\ref{fig:1d_ben_graphserr_adapt}(b) shows here again that the neural net training consistently succeeds.

\section{Conclusion and outlook}\label{sec:conclusion}
We have studied the domain adaptation in our method from~\cite{crisan2022application} on the example of the Benes filter.
We observed that the domain adapted method was more effective in resolving the bimodality than the non-domain adapted one. However, this came at the cost of a linear trend in the error. A possible direction for future work would thus be to investigate the optimal domain size more closely, in order to mitigate the error trend, and make full use of the increased resolution from the domain adaptation. This is subject of future research in connection with more general domain adaptation methods than the one employed here, which is specific to the Benes filter.

As already noted in the previous work~\cite{crisan2022application},
the possibility for \emph{transfer learning} in our method should be explored.

A long-term goal in the development of neural network based numerical methods must of course be the rigorous error analysis, which remains a challenging task.

\clearpage
\setlength{\bibitemsep}{2.5 pt}

\printbibliography

\end{document}